# A adaptive block based integrated LDP, GLCM, and Morphological features for Face Recognition

Arindam Kar[1], Debotosh Bhattacharjee[2], Dipak Kumar Basu[2*], Mita Nasipuri[2], Mahantapas Kundu[2]
[1] Indian Statistical Institute, Kolkata-700108, India
[2] Department of Computer Science and Engineering, Jadavpur University, Kolkata- 700032, India
* AICTE Emeritus Fellow
Email: {kgparindamkar@gmail.com, debotosh@indiatimes.com, dipakkbasu@gmail.com, m.nasipuri@cse.jdvu.ac.in, mkundu@cse.jdvu.ac.in}

*Abstract*— This paper proposes a technique for automatic face recognition using integrated multiple feature sets extracted from the significant blocks of a gradient image. We discuss about the use of novel morphological, local directional pattern (LDP) and gray-level co-occurrence matrix GLCM based feature extraction technique to recognize human faces. Firstly, the new morphological features i.e., features based on number of runs of pixels in four directions (N,NE,E,NW) are extracted, together with the GLCM based statistical features and LDP features that are less sensitive to the noise and non-monotonic illumination changes, are extracted from the significant blocks of the gradient image. Then these features are concatenated together. We integrate the above mentioned methods to take full advantage of the three approaches. Extraction of the significant blocks from the absolute gradient image and hence from the original image to extract pertinent information with the idea of dimension reduction forms the basis of the work. The efficiency of our method is demonstrated by the experiment on 1100 images from the FRAV2D face database, 2200 images from the FERET database, where the images vary in pose, expression, illumination and scale and 400 images from the ORL face database, where the images slightly vary in pose. Our method has shown 90.3%, 93% and 98.75% recognition accuracy for the FRAV2D, FERET and the ORL database respectively.

*Index Terms*—Face recognition, Feature extraction, Morphological operations, local directional pattern (LDP) Analysis, gray-level co-occurrence matrix (GLCM), significant blocks, gradient image .

## I. INTRODUCTION

Face recognition has attracted significant attention because of its wide range of applications [1]. One problem of face recognition is the fact that different faces could seem very similar; therefore, a discriminating task is required, which is not so easy. Even though human beings can detect and identify faces with no effort, building an automated system that accomplishes such objectives is very challenging. The challenges are even more profound when there are large variations due to illumination conditions, change in pose, facial expression etc. Of all image analysis of human face, feature extraction is of immense importance. The motivation of feature based method is due to the representation of the facial images in a very compact way, i.e., dimensionality reduction and hence lowering the memory needs.

On the other hand, when we analyze the same face, many characteristics may have changed. These variations might be because of changes in the different parameters like illumination, variability in facial expressions, the presence of accessories (glasses, beards, etc); poses, age, finally background [2], [3],[4]. To combine image information (such as color, gradient, movement, and invariant moments) and knowledge of the face (such as symmetry of the face, statistical relative position among the mouth, the nose and the eyes) is an important trend. A variety of techniques have been used for measuring texture and transformation to other spaces such as eigenspace [5], fisherspace and Linear Discriminant Analysis (LDA) [6], Principal component Analysis (PCA)[7], Fractals , Gabor filters, variations of wavelet transform [8]. Many researchers have used the gray-level co-occurrence matrix [9] for the extraction of features to be used in texture classification. Gelzinis et al. [10] presented a new approach to exploiting information available in the co-occurrence matrices computed for different distance parameter values. The identification of specific textures in an image is achieved primarily by modeling texture as a two-dimensional gray level variation. This two dimensional array is called as Gray Level Co-occurrence Matrix (GLCM).

The idea behind this paper is to integrate the GLCM based features together with a new morphological approach based features [11], and the local directional pattern (LDP) analysis based features [12], and takes the full advantage of the above methods from only the significant blocks of the gradient image rather than the whole image which reduces the computational complexity As LDP is more stable than local binary pattern (LBP) [13], in presence of noise so LDP is preferred over LBP.

In this paper the block based facial features are compared locally, instead of using a general structure, hence it allows us to make a decision from the parts of the face. For example, when there are sunglasses, the algorithm compares faces in terms of mouth, nose and any other features rather than eyes.
Moreover as a result of using local distinct features, instead of a face template or local features bounded by a graph, proposed method gives a high performance result on occluded cases.

Here firstly the changes in contrast within the image are determined by calculating the gradient of the input image. After calculating the gradient image, the mean and median of the gradient image is calculated and multiple of the average of mean and median is used to calculate the threshold value, which in turn gives a binary image. In this paper the gradient image is used for recognition instead of the original image. From this gradient image m blocks of size u×v pixels are selected randomly which are greater than the average of the mean and median of the absolute gradient image.

To the best of the literature we have surveyed, no one has attempted before to implement this concept of applying this method on the significant blocks of the absolute binary gradient image. The idea is simple and straight forward. For each face image, a 56 dimensional feature vector is extracted from the LDP analysis, Secondly; a 4 dimensional feature vector is extracted by applying the morphological operations in four directions. Then a GLCM based 12 dimensional feature vector is extracted which contains the statistical information. Finally these extracted feature vectors are concatenated together to obtain an integrated 72 dimensional feature vector which is then used for classification.

The remainder of this paper is organized as follows: section II describes the derivation of most informative blocks from the eight directional gradient images of the facial images. Section III details the extraction of GLCM, LDP and morphological features from the extracted blocks. Section IV deals the similarity measure and classification rule. In Section V and VI we assesses the performance of the proposed method on the face recognition task by applying it on the FERET [14], FRAV2D [15] and the ORL face [16] databases and finally by comparing with some of the most popular face recognition schemes we conclude our paper.

## II. FACE IDENTIFICATION

### A. Absolute Binary Gadient Image

In this section the gradient image $I_G$ of an image (I) is calculated by taking summation over the absolute of the change of pixels in all the eight particular directions i.e., (north (N), northeast (NE), east (E), southeast (SE), south (S), southwest (SW), west (W) and northwest (NW)). The absolute binary gradient image $I_{AG}$ is calculated by considering the pixels of $I_G$ those are greater than the average of the mean and median of $I_G$. The original image and its corresponding binary absolute gradient image are shown in Fig. 1 (a) and Fig. 1 (b) respectively.

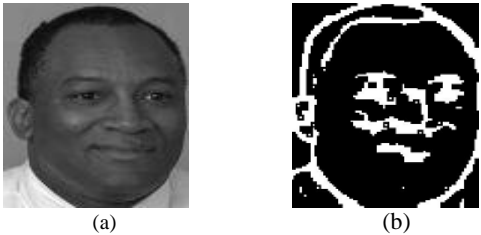

Fig 1.(a) Original image, (b) Binary absolute gradient image

### A. Image enhancement and Binarization

As a gradient image can have values not in the range of [0,255]. Thus image enhancement becomes necessary for the gradient image. This is done by subtracting the minimum value of the gradient image from the gradient image of the image. Then all the pixels in the modified image are scaled to the range [0,255]. After this enhancement, posterization procedure is used to eliminate the background influence. Posterization occurs when an image apparent bit depth has been decreased so much that it has a visual impact. Let I(x,y) be a gray scale image, and $n_L$ the number of levels considered for posterization. The image enhancement and posterization is done as follows:

$I = I - minimum(I)$ ;

$I_e = round\left(\frac{I}{maximum\ (I)} \cdot 255\right);$

$I_p = \sum_{x,y} round\left(\frac{I_e(x,y)}{256} n_L\right) \cdot \frac{256}{n_L}$

From the above figure 1(b) it is clear that the white portions of the absolute gradient image where the gradient is very high and those are the region of interest i.e., we are interested to extract the circular regions that contains the important facial features like nose, mouth, eye etc.

### B. Extraction of Significant Blocks

The step for extracting significant blocks from the image is given below.
Blocks are generated randomly within the absolute gradient image $I_{AG}$ and are selected if
 a) The average count of white pixels i.e., 1 in the block is greater than M say, where M is the average of mean and median of the pixel values of the overall binary gradient image.
b) If there is any overlap between a selected block and a new block, then that particular block is selected for which the count of white pixels is more.
c) Select that same position block from the original image (I). The $k^{th}$ extracted block for the $i^{th}$ training image is defined as $B_{i,j} = \{(x_k, y_k), B_{i,j}(x_k, y_k)\}$. The first two components represent the starting location of the extracted block . The first two components of feature are very important during matching (comparison) process. The remaining components are the elements of the block.
d)This process of random block generation and extraction is repeated for K times (say K=100000). The most significant blocks are selected from these randomly generated blocks in such a way that the blocks capture the important facial features like nose, mouth, eye etc and are in the range of 10-15% of the total blocks non overlapping blocks in which the image is divided. The image is divided into 90 non overlapping blocks: 9 vertical and 10 horizontal blocks. The proposed block division has been adjusted experimentally.

## II. FEATURE EXTRACTION

### A. GLCM based features

One of the simplest approaches for describing texture is to use statistical moments of the intensity histogram of an image [17]. Using a statistical approach such as co-occurrence matrix will help to provide valuable information about the relative position of the neighbouring pixels in an image.

The GLCM or Gray Level Co-occurrence Matrix is a tabulation of how often different combinations of pixel brightness values (grey levels) occur in an image. It was developed by Haralick. Mathematically, a co-occurrence matrix C is defined over an n × m image I, parameterized by an offset (Δx, Δy), as:

$$G(i,j) = \sum_{p=1}^{n}\sum_{q=1}^{m}\begin{cases}1, if\ I(x,y)=i\ \&\ I(x+\Delta x, y+\Delta y)=j\\ 0,\qquad otherwise\end{cases}$$

where the (Δx, Δy), is specifying the distance between the pixel-of-interest and its neighbour. Note that the offset (Δx, Δy), parameterization makes the co-occurrence matrix sensitive to rotation. Choosing the offset vector, such that a rotation of the image not equal to 180 degrees, will result in a different co-occurrence matrix for the same (rotated) image. This can be avoided by forming the co-occurrence matrix using a set of offsets sweeping through 180 degrees at the same distance parameter Δ to achieve a degree of rotational invariance (i.e. [0 Δ] for 0°: G horizontal, [-Δ Δ] for 45°: G right diagonal, [-Δ 0] for 90°: G vertical, and [-Δ -Δ] for 135°: G left diagonal.

Fig. 2 shows the generation of four co-occurrence matrices using $N_g = 5$ levels and offsets {[0 1], [-1 1], [-1 0], [-1 -1]} defined as one neighboring pixel in the possible four directions.

Fig. 2 Cooccurrence matrix generation for $N_g$=5 levels and four different offsets: $G_{LD}$ (135°), $G_H$ (0°), $G_{RD}$ (45°), $G_V$ (90°).

Find four GLCM for a block, one for each direction. $0°, 45°, 90°, 135°$ and the resulting GLCM is taken as the average of these four matrices, which is given by :

$$G = \frac{G_H + G_V + G_{RD} + G_{LD}}{4}.$$

The GLCM is directly used after converting it to a column vector. However, for 8-bit gray level representation the size of this vector is 256×256, which corresponds to a 65536 dimensional vector. The dimension of this vector can be reduced by reducing the number of gray levels, Ng, of the image. For example, if Ng is reduced to 16, the GLCM will be 256 (16×16) dimensional vector. From this average GLCM matrix, a vector of these 4 statistical features are calculated as follows :

a) Energy = $\sum_{i,j} G(i,j)$
b) Contrast = $\sum_{k=0}^{m-1} k^2 \sum_{|i-j|=k} G(i,j)$
c) Correlation = $\sum_{i,j} \frac{(i-\mu)(j-\mu)G(i,j)}{\sigma^2}$
d) Homogeneity = $\sum_{i,j} \frac{G(i,j)}{1+|i-j|}$

where $\mu = \sum_{i,j} iG(i,j)$, (weighted pixel average)
σ =(weighted pixel variance)= $\sum_{ij}(1-\mu)^2 . G(i,j)$

The step for extracting GLCM based feature from the significant blocks is given below.

For each block $B_i$, i=1, 2,…n the 4 GLCM based features are calculated say $f_i = (f_{i1}, ...., f_{i4})$. The feature vector $F = (f_1, ..., f_n)^T$. Hence we obtain a feature vector of size 4 × n where n is the number of blocks. For further precision:

b) We also consider GLCM for distance = 1, 2 and 3 units. This makes the feature vector size as 3 × (4n) or 12 × n. Despite this slight increase in size of the feature vector, the classifier performance is largely improved.

### B. Edge based features

Edges of an image give us information about some structural features of the image. As a morphological operator contains inherent properties to capture shape and structure of an image, we find the number of runs of edges in a particular direction. $0°, 45°, 90°, 135°$. Then find out the number of runs of change of pixel in a particular direction for the significant blocks.

$M_0 = \begin{pmatrix} c & 1 \end{pmatrix}$, $M_{90} = \begin{pmatrix} c \\ 1 \end{pmatrix}$, $M_{45} = \begin{pmatrix} c & 0 \\ 0 & 1 \end{pmatrix}$, and

$M_{135} = \begin{pmatrix} 0 & 1 \\ c & 0 \end{pmatrix}$, where c denotes the centre of the structuring element. The morphological erosion of any block with $M_0$ gives the number of runs in the horizontal direction, similarly $M_{90}$, $M_{45}$, $M_{135}$ give the runs in the other three directions.

The step for extracting morphological feature from the significant blocks is given below:

a. Find the runs in four directions for each block using morphological erosion. Let them be termed as $R_i = (r_1, ..., r_4)$.

b. The feature vector is $F = (R_1, ..., R_n)^T$.

## C. LDP based features

The Local Directional Pattern (LDP), proposed in [12], computes for each pixel of the image the edge response value in different directions and uses these values to encode the image. The LDP includes similar information like that of LBP provides but produces more stable pattern in presence of noise and non monotonic illumination changes since gradients are more stable than gray level. Concisely, [12] calculates eight directional edge responses value for each pixel using Kirsch masks in eight different orientations. These masks are shown in Fig. 3. Applying eight masks, we obtain eight edge response values $m_0, m_1, \ldots, m_7$.

$$\begin{pmatrix} -3 & -3 & 5 \\ -3 & 0 & 5 \\ -3 & -3 & 5 \end{pmatrix} \quad \begin{pmatrix} -3 & 5 & 5 \\ -3 & 0 & 5 \\ -3 & -3 & -3 \end{pmatrix} \quad \begin{pmatrix} 5 & 5 & 5 \\ -3 & 0 & -3 \\ -3 & -3 & -3 \end{pmatrix} \quad \begin{pmatrix} 5 & 5 & -3 \\ 5 & 0 & -3 \\ -3 & -3 & -3 \end{pmatrix}$$

$M_0$ (East)    $M_1$ (North-East)    $M_2$ (North)    $M_3$ (North West)

$$\begin{pmatrix} 5 & -3 & -3 \\ 5 & 0 & -3 \\ 5 & -3 & -3 \end{pmatrix} \quad \begin{pmatrix} -3 & -3 & -3 \\ 5 & 0 & -3 \\ 5 & 5 & -3 \end{pmatrix} \quad \begin{pmatrix} -3 & -3 & -3 \\ -3 & 0 & -3 \\ 5 & 5 & 5 \end{pmatrix} \quad \begin{pmatrix} -3 & -3 & -3 \\ -3 & 0 & 5 \\ -3 & 5 & 5 \end{pmatrix}$$

$M_4$ (West)    $M_5$ (South – West)    $M_6$ (South)    $M_7$ (South-East)

Fig 3. Kirsch edge response masks in eight directions

The calculated values $m_0, m_1, \ldots, m_7$ represent the edge significance in its respective direction. The presence of edges and its orientation will determine the highest response values in each particular direction. The LDP code the k most prominent directions. Hence, the top k values $|m_j|$ are set to 1 and the other 5 values are set to 0. For instance, Fig. 4 shows an original image and the corresponding image after adding Gaussian white noise. After addition of noise, 5th bit of LBP changed from 1 to 0, thus LBP pattern changed from uniform to a non-uniform code. Since gradients are more stable than gray value, LDP pattern provides the same pattern value even presence of that noise and non-monotonic illumination changes.

| Mask index | $m_7$ | $m_6$ | $m_8$ | $m_4$ | $m_2$ | $m_2$ | $m_1$ | $m_0$ |
|---|---|---|---|---|---|---|---|---|
| Mask Value | 161 | 97 | 161 | 537 | 313 | 97 | -503 | -393 |
| Rank | 6 | 7 | 5 | 1 | 4 | 8 | 2 | 3 |
| Code Bit | 0 | 0 | 0 | 1 | 0 | 0 | 1 | 1 |
| LDP Code | | | | 19 | | | | |

(with 3x3 grid: 85 32 26 / 50 50 10 / 60 38 45 →)

Fig. 4. LPD Code generation

After encoding an image with the LDP operator we get an encoded image $I_L$. Here k = 3 is used which generates 56 distinct values in the encoded image. So histogram H of this LDP labeled image $I_L(x, y)$ is a 56 bin histogram and can be defined as
$H = \sum_{x,y} P(I_L(x, y) = C_i)$,
$c_i = i^{th}$ *LPD pattern* $(0 \leq i < 56)$
where P(A) = 1 if A is true, else 0.

The step for extracting features based on LDP analysis from the extracted significant blocks of the image is given below.

1. Calculate the histograms of each block taking into account that the LDP is obtained by coding the three predominant directions (there are just three ones of the 8 bits code), so there are just 56 possible values{7, 11, 13, 14, ... , 200, 208, 224}. Therefore, a 56 bin histogram of each block is calculated $\{H_{LDP}^i\}_{i=1}^n$ being the bin centers on the possible values.

The histograms of each block are concatenated to obtain a feature vector size as a 56×n or 56n. The histogram acts as the feature vector of dimension 56n.

Thus from LDP analysis a feature vector with 56n components are obtained; from morphological operation i.e. feature based on the number of runs of pixels in four directions (N, NE, E, NW) a feature vector of 4n component is obtained and from GLCM for distance = 1, 2 &3 unit the haralik features are extracted i.e. a feature vector of 12×n components are obtained. These features are concatenated to obtain a final feature vector ($F_n$) having 72×n components.

IV. SIMILARITY MEASURE AND CLASSIFICATION

The dissimilarity measure between the $i^{th}$ testing and $j^{th}$ training image is measured as:

$$D(i,j) = \frac{1}{B_i} \sum_{k=1}^{B_i} \left[ \max_{1 \leq l \leq B_j} \{I(k,l) \cdot \aleph^2(f_{ik}, f_{jl})\} \right]$$

where $B_i$ = number of blocks in the $i^{th}$ testing image, $B_j$ = number of blocks in the $j^{th}$ training image. Here $I(k, l)$ is defined in such a way that blocks of the training image which are not close enough to a block of the test image in terms of location are discarded, and the extracted features of those training blocks which are in the neighbourhood of the testing blocks are only compared for finding the similarity measure i.e,

$$I(k,l) = \begin{cases} 1 & if \sqrt{(x_k - x_l)^2 + (y_k - y_l)^2} < th1 \\ 0 & otherwise \end{cases}$$

Here $th1$ is the threshold; in this experiment the threshold is taken as the block size i.e. the approximate radius of the area that contains the eye. This technique helps to avoid the matching of a feature of a block located around the eye with a point of a training facial image that is located around the mouth.

Also $\aleph^2(f_1, f_2)$ is the Chi-square [18] dissimilarity measure between two feature vector $f_1$ and $f_2$, which is defined as:

$$\aleph^2(f_1, f_2) = \sum_{m=1}^{72} \frac{(f_{1m} - f_{2m})^2}{(f_{1m} + f_{2m})}$$

where m is the number of features in the block.
Finally, the i$^{th}$ testing image is classified to the k$^{th}$ training image class
if $D(i,j) = \min_j D(i,j)$.

## V. EXPERIMENT

The feasibility of the proposed method has been successfully tested on face recognition using three databases, a) the whole ORL facial database, b) the FRAV2D database containing 1100 frontal face images corresponding to 100 subjects, c) The FERET database, containing 2200 frontal face images corresponding to 200 subjects, which are acquired under variable illumination and facial expression. The effectiveness of the proposed method is shown in terms of both absolute performance indices and comparative performance against some popular face recognition schemes such as the PCA, LDA, Gabor wavelets (GW), the LBP, direct GLCM, direct LDP, edge based features and their different combinations.

### A. Experiment on the ORL database

The whole ORL database, is considered here, each image is scaled to $92 \times 112$ with 256 gray levels.

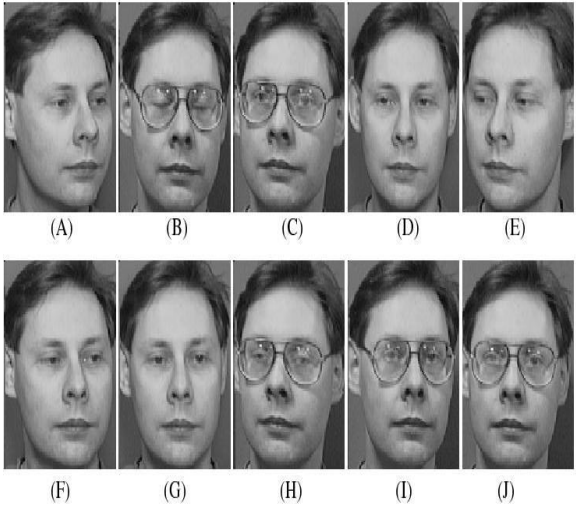

Fig. 5. Demonstration images of an individual from the ORL database

### B. Experiment on the FRAV2D database

The FRAV2D face database, employed in the experiment, consists of 1100 colour face images of 100 individuals, 11 images of each individual are taken, including frontal views of faces with different facial expressions, under different lighting conditions. All colour images are transformed into gray images and scaled to 92×112. Fig. 6 shows all samples of one individual. The details of the images are as follows: (A) regular facial status; (B) and (C) are images with a 15° turn with respect to the camera axis; (D) and (E) are images with a 30° turn with respect to the camera axis; (F) and (G) are images with gestures; (H) and (I) are images with occluded face features; (J) and (K) are images with change of illumination.

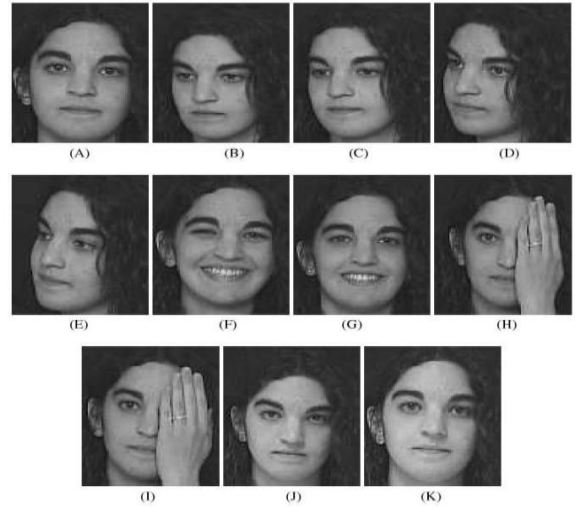

Fig 6. Demonstration images of an individual from the FRAV2D database

### C. Experiment on the FERET database

The FERET database, employed in the experiment here, contains 2,200 facial images corresponding to 200 individuals with each individual contributing 11 images. The images in this database were captured under various illuminations and display, a variety of facial expressions and poses. As the images include the background and the body chest region, so each image is cropped to exclude those, and then scaled to $92 \times 112$. Fig. 7 shows all samples of one individual. The details of the images are as follows: (A) regular facial status; (B) +15° pose angle; (C) -15° pose angle; (D) +25° pose angle; (E) -25° pose angle; (F) +40° angle; (G) -40° pose angle; (H) +60° pose angle; (I) -60° pose angle; (J) alternative expression; (K) different illumination.

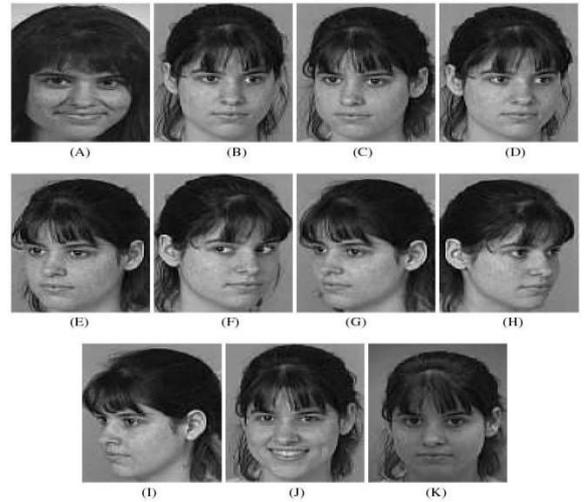

Fig. 7. Demonstration images of an individual from the FERET database

### D. Specificity and Sensitivity measure for the FRAV2D, FERET and the ORL dataset:

To measure the sensitivity and specificity [19, 20] the dataset from the ORL, FRAV2D, and FERET database are prepared in the following manner. From the ORL database for each individual a single class is constituted, i.e., 40 classes are constructed with 15 images in each class. Out of these 15 images in each class, 10 images are of a particular individual, and 5 images are of other individuals. In the FRAV2D database also for each individual a single class is constituted with 18 images in it. Out of the 18 images in each class, 11 images are of a particular individual, and 7 images are of other individuals taken by permutation. Thus a total 100 class is obtained, containing 18 images in it, from the dataset of 1100 images of 100 individuals. Similarly 200 classes are obtained for the FERET dataset, each class having 18 images in it. Out of the 18 images in each class, 11 images are of a particular individual, and 7 images are of other individuals taken by permutation Using these datasets the true positive ($T_P$); false positive ($F_P$); true negative ($T_N$); false negative ($F_N$); are being measured, and hence the specificity and sensitivity. For the ORL dataset, only the first 2 images (A-B), of a particular individual are selected as training samples and the remaining images of that particular individual are used as positive testing samples. For the FERET and FRAV2D dataset the first 3 images (A-C), of a particular individual are selected as training samples and the remaining images of that particular individual are used as positive testing samples. The negative testing is done using the images of the other individuals for all the databases.

Table I: Specificity and Sensitivity measure of the ORL, FERET and FRAV2D database:

| Total no. of classes=40, Total no. of images= 600 | | | |
|---|---|---|---|
| ORL | | Individual belonging to a particular class | |
| | | Using first 2 images of an individual as training images | |
| ORL test | Positive | $T_P$ =316 | $F_P$ =0 |
| | Negative | $F_N$ =4 | $T_N$ =200 |
| | | Sensitivity = $T_P$ / ($T_P + F_N$) ≈ 99% | Specificity = $T_N$ / ($F_P + T_N$)=100% |
| Total no. of classes=100, Total no. of images= 1800 | | | |
| FRAV2D | | Individual belonging to a particular class | |
| | | Using first 3 images of an individual as training images | |
| | | Positive | Negative |
| FRAV2D test | Positive | $T_P$ =744 | $F_P$ =7 |
| | Negative | $F_N$ =56 | $T_N$ =693 |
| | | Sensitivity = $T_P$ / ($T_P + F_N$) ≈ 93% | Specificity = $T_N$ / ($F_P + T_N$) ≈99.3% |
| Total no. of classes=200, Total no. of images= 3600 | | | |
| FERET | | Individual belonging to a particular class | |
| | | Using first 3 images of an individual as training images | |
| | | Positive | Negative |
| FERET test | Positive | $T_P$ = 1445 | $F_P$ =14 |
| | Negative | $F_N$ = 155 | $T_N$ =1386 |
| | | Sensitivity = $T_P$ / ($T_P + F_N$) ≈ 90.3% | Specificity = $T_N$ / ($F_P + T_N$)=99% |

For **ORL database** considering the first 2 images (A-B) of a particular individual for training the achieved rates are:
**False positive rate** = FP / (FP + TN) = 1 − Specificity =0%
**False negative rate** = FN/(TP+ FN) = 1 − Sensitivity=**1**%
**Accuracy** = ($T_P+T_N$)/($T_P+T_N+F_P+F_N$) = **99.5.**

Thus for **FRAV2D db** considering the first 3 images (A-C) of a particular individual for training the achieved rates are:
**False positive rate** = FP / (FP + TN) = 1 − Specificity =.7%
**False negative rate** = FN / (TP + FN) = 1− Sensitivity=**7**%
**Accuracy** = ($T_P+T_N$)/($T_P+T_N+F_P+F_N$) ≈ **96.2.**

For **FERET database** considering the first 3 images (A-C) of a particular individual for training the achieved rates are:
**False positive rate** = FP / (FP + TN) = 1 − Specificity =1%
**False negative rate** = FN/(TP+ FN) = 1 − Sensitivity=**10.7**%
**Accuracy** = ($T_P+T_N$)/($T_P+T_N+F_P+F_N$) ≈ **94.2.**

## VI. RESULTS

Experimental results indicate that a) the extracted features by the GLCM, LDP analysis and the morphological operations, from the proposed significant blocks of the absolute gradient image integrated together used for classification using the $\chi^2$ distance similarity measure achieved a verification rate which is as good as any previously used methods like PCA, LDA, Gabor wavelets, LBP, LDP .It is also seen that LDP based feature extraction technique achieves the best accuracy, when combined with GLCM based features and edge based features; b) During simulations, it is observed that the locations of significant blocks , found from the absolute binary gradient image of the face image, can give small deviations between different conditions (expression, illumination, having glasses or not, rotation, etc.), for the same individual. Therefore, an exact measurement of corresponding distances is not possible unlike the geometrical feature based methods; c) The computational time is significantly reduced by using only few significant blocks of the image, although the accuracy is not compromised; and finally d) the feature are automatically extracted using the local characteristics of an individual face, in order to decrease the effect of occluded features. In Table II, shows the current upper bound performances on the FERET, FRAV2D and ORL facial database, which reflects that our proposed method achieves higher performance results. Thus the proposed algorithm deals with two of these problems, namely occlusion and illumination changes.

Experimental on all the three databases taking first **two** images (A-B) as **training images** for the ORL database and **three** images (A-C) as **training images** for FERET and FRAV2D database. The **remaining** images are considered testing images for all the databases. The recognition rate of the proposed method is compared with some other well known methods like Gabor wavelets, LBP, PCA, and LDA. Then recognition performance is evaluated using only GLCM, LDP, and edge based features techniques and wit their possible combination, using the distance measure $\delta_\aleph$ and is shown in Table II.

Table II. Performance results of well known face recognition algorithms together with the proposed method on FERET, ORL and FRAV2D respectively with the use of $\chi^2$ similarity measure.

| Database | Feature Combination | Highest Recognition Accuracy |
|---|---|---|
| **FERET, ORL, FRAV2D** | **GLCM+LDP+EDGE** | **90.3, 98.75, 93** |
| FERET,ORL, FRAV2D | LDP+EDGE | 88.4, 96.6, 91.7 |
| FERET,ORL, FRAV2D | GLCM+LDP | 86.75, 92.7, 90.4 |
| FERET,ORL, FRAV2D | GLCM +EDGE | 82.9, 90.5, 88.75 |
| FERET,ORL, FRAV2D | LDP | 82.75, 88.5, 87.5 |
| FERET,ORL, FRAV2D | LBP | 81.8, 87.8, 85.6 |
| FERET,ORL, FRAV2D | Gabor Wavelets | 79.5, 87.56, 82.4 |
| FERET,ORL, FRAV2D | EDGE | 78.2, 85, 81.8 |
| FERET,ORL, FRAV2D | GLCM | 76, 80.5, 78.75 |
| FERET,ORL, FRAV2D | LDA | 74.67, 86.67, 79.4 |
| FERET,ORL, FRAV2D | PCA | 71.5, 82.86, 76 |

## CONCLUSION

The proposed technique as presented here obtains gradient image of the original image in eight directions and takes its absolute value. From these gradient images only the informative significant blocks are extracted with our new extraction technique, and are used for enhanced face recognition purpose. In this approach as only the significant blocks are used instead of the whole image, makes the method computationally efficient and suitable for real time application. Here as the facial features are selected only from the significant blocks. Also in this approach the facial features are compared locally instead of a general structure, and so it allows us to make a decision from the different parts of a face. Thus it performs better in presence of occlusions. Since the LDP, features based on the number of runs on the edge and GLCM based features are extracted from only the significant blocks and concatenated for enhanced face recognition The proposed method is also robust to noise, non-monotonic illumination variation, and expression changes as a property of the LDP analysis based features. By integrating the two complementary approaches (LDP and GLCM), with the edge based approach, and applying only on the most significant blocks not only makes the new method efficient, but also makes the new method achieves a superior performance than the approaches alone.

## ACKNOWLEDGEMENT

Authors are thankful to a major project entitled "Design and Development of Facial Thermogram Technology for Biometric Security System," funded by University Grants Commission (UGC),India and "DST-PURSE Programme" and CMATER and SRUVM project at Department of Computer Science and Engineering, Jadavpur University, Kolkata - 700 032 India for providing necessary infrastructure to conduct experiments relating to this work.